\documentclass[conference]{IEEEtran}
\IEEEoverridecommandlockouts
\usepackage{cite}
\usepackage{multirow}
\let\OLDthebibliography\thebibliography
\renewcommand\thebibliography[1]{
  \OLDthebibliography{#1}
  \setlength{\parskip}{0pt}
  \setlength{\itemsep}{0pt plus 0.3ex}
}
\usepackage{subfigure}
\usepackage{enumitem}
\usepackage{hyperref}
\usepackage{bm}
\usepackage{balance}
\usepackage{marvosym}
\usepackage{url}
\usepackage{amsmath,amssymb,amsfonts}
\usepackage{algorithmic}
\usepackage{graphicx}
\usepackage{textcomp}
\usepackage{xcolor}
\begin{document}

\title{Improving Audio-Visual Speech Recognition by Lip-Subword Correlation Based Visual Pre-training and Cross-Modal Fusion Encoder}

\author{
\IEEEauthorblockN{1\textsuperscript{st} Yusheng Dai}
\IEEEauthorblockA{
\textit{University of Science and Technology} \\ \textit{of China}\\
Hefei, China \\}      
\and
\IEEEauthorblockN{2\textsuperscript{nd} Hang Chen}
\IEEEauthorblockA{
\textit{University of Science and Technology} \\ \textit{of China}\\
Hefei, China \\}
\and
\IEEEauthorblockN{3\textsuperscript{rd} Jun Du\textsuperscript{*}\thanks{\textsuperscript{*}corresponding author}}
\IEEEauthorblockA{
\textit{University of Science and Technology} \\ \textit{of China}\\
Hefei, China \\}
\and
\IEEEauthorblockN{4\textsuperscript{th} Xiaofei Ding}
\hspace{100pt}
\IEEEauthorblockA{\textit{Alibaba Group}\\
Hangzhou, China \\}
\and
\IEEEauthorblockN{5\textsuperscript{th} Ning Ding}
\hspace{100pt}
\IEEEauthorblockA{\textit{Alibaba Group}\\
Hangzhou, China}
\and
\IEEEauthorblockN{6\textsuperscript{th} Feijun Jiang}
\hspace{100pt}
\IEEEauthorblockA{\textit{Alibaba Group}\\
Hangzhou, China}
\and
\IEEEauthorblockN{7\textsuperscript{th} Chin-Hui Lee}
\IEEEauthorblockA{\textit{Georgia Institute of Technology}\\
Atlanta, USA}
}

\maketitle

\begin{abstract}
In recent research, slight performance improvement is observed from automatic speech recognition systems to audio-visual
speech recognition systems in end-to-end frameworks with low-quality videos. Unmatching convergence rates and specialized input representations between audio-visual modalities are considered to cause the problem. In this paper, we propose two novel techniques to improve audio-visual speech recognition (AVSR) under a pre-training and fine-tuning training framework. First, we explore the correlation between lip shapes and syllable-level subword units in Mandarin through a frame-level subword unit classification task with visual streams as input. The fine-grained subword labels guide the network to capture temporal relationships between lip shapes and result in an accurate alignment between video and audio streams. Next, we propose an audio-guided Cross-Modal Fusion Encoder (CMFE) to utilize main training parameters for multiple cross-modal attention layers to make full use of modality complementarity. Experiments on the MISP2021-AVSR data set show the effectiveness of the two proposed techniques. Together, using only a relatively small amount of training data, the final system achieves better performances than state-of-the-art systems with more complex front-ends and back-ends. The code is released at\footnote{https://github.com/mispchallenge/MISP-ICME-AVSR}.
\end{abstract}


\section{Introduction}
\label{sec:intro}
Audio-visual speech recognition (AVSR) is a multi-modality application motivated by the bi-modal nature of perception in speech communication between humans~\cite{besle2004bimodal}. It utilizes lip movement as a complementary modality to improve the performance of automatic speech recognition (ASR). In early research, handcrafted lip features were commonly extracted and added to hybrid ASR systems~\cite{  1326155,4782036,1230212,dupont2000audio}. Recently, end-to-end AVSR systems have achieved great success due to the simplicity of end-to-end ASR system designs and an availability of a large number of public audio-visual databases~\cite{8099850,chung2016lip,chen2022audio,cooke2006audio,7163155,zhao2009lipreading,liu2020robust,7918453,10.1145/3338533.3366579}. Although end-to-end AVSR systems have shown their simplicity and effectiveness on many benchmarks \cite{ma2021end,xu2022channel,afouras2018deep,zhou2019modality,yu2020audio,makino2019recurrent}, they are still far from common use. One hard piece of evidence comes from the recent MISP2021 Challenge~\cite{chen2022first}.

As the largest Mandarin audio-visual corpus until now, MISP2021-AVSR is recorded in TV rooms of home environments with multiple groups chatting simultaneously. Multiple microphone arrays and cameras are used to collect far/middle/near-field audios and far/middle-field videos. In the evaluation stage, submitted systems are restricted to utilizing far-field audio and videos for the AVSR task. According to the reports from top-ranked teams ~\cite{xu2022channel,wangxiaomi,wang2022sjtu}, low-quality far-field videos could only slightly improve the AVSR performance over ASR systems under a common end-to-end framework. The performance degradation from a uni-modal network to a multi-modal network is also observed in~\cite{wang2020makes}. Compared to the uni-modal model, it is challenging to learn an extensive integrated neural network due to unmatched convergence rates and specialized input representations between two modalities~\cite{nagrani2021attention,wang2020makes}. Pre-training techniques ~\cite{ma2021end,Zhang_2019_ICCV,pan2022leveraging} are expected to alleviate the problem which decouples the one-pass end-to-end training framework in two stages. Uni-modal networks are first pre-trained and later integrated into a fusion model following unified fine-tuning. This divide-and-conquer strategy could effectively mitigate variations in learning dynamics between modalities and promote their interactions.

 \begin{figure*}[htbp]
 \vspace{-8mm}
 \centering
 \includegraphics[width=1.94\columnwidth]{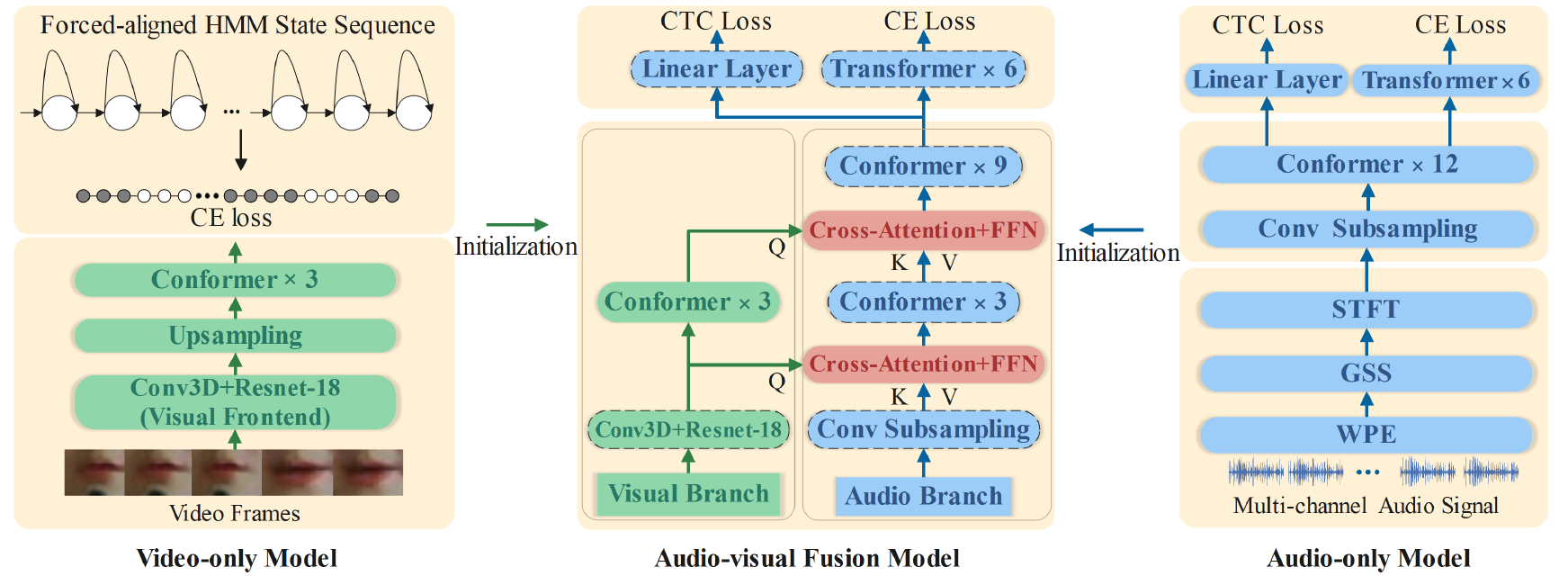}
    \caption{Overall training framework of our AVSR system}
    \label{fig:framework}
\end{figure*}

A crucial aspect of the decoupled training framework is how to pre-train the visual frontend. A simple practice comes from~\cite{xu2022channel,wangxiaomi}, which directly uses a pre-trained visual frontend~\cite{martinez2020lipreading} as a frozen visual embedding extractor. This approach gives only a small improvement due to domain shifts across the source and target domains. For most studies, researchers pre-train the visual frontend on an isolated word recognition task ~\cite{ma2021end,petridis2018audio,Zhang_2019_ICCV} and then fine-tune it with the AVSR model. However, these pre-training methods depend on word-level lipreading data sets that are challenging to collect on a large scale. A recent study~\cite{pan2022leveraging} leverages self-supervised learning on large-scale unlabeled data sets for AVSR. Although these pre-training methods improve the AVSR system performances to a certain extent, a large amount of extra labeled/unlabeled data is used.

In this paper, we propose a subword-correlated visual pre-training technique that does not need extra data or manually-labeled word boundaries. We train a set of hidden Markov models with Gaussian mixture model (GMM-HMMs) on far-field audio to produce frame-level alignment labels and pre-train the visual frontend by identifying each visual frame's corresponding syllable-related HMM states. Compared to the pre-training method based on end-to-end continuous lipreading, our method explicitly offers syllable boundaries to establish a direct frame-level mapping from lip shapes to syllables in Mandarin. These fine-grained alignment labels guide the network to focus on learning visual feature extraction of low-quality videos. On the other hand, this pre-training method could be viewed as a cross-modal conversion process that accepts video frames as inputs and generates acoustic subword sequences. It is helpful to explore potential acoustic information from lip movements and contributes to a good adaptation process with the audio stream in the fusion stage.\\
\indent In the fusion stage of decoupled training, the initialized audio and visual branches already have the fundamental ability to extract uni-modal representations. Based on the straightforward assumption that the audio modality contains more linguistic information essential for ASR tasks. We propose a novel CMFE block in which the audio modality dominates and more training parameters of the network are used for modality fusion modeling. As for the modality fusion structures, motivated by the decoder architecture of the vanilla transformer~\cite{vaswani2017attention}, the layer-wise cross-attention is designed in different layers to make full use of modality complementarity.\\
\indent In summary, for this paper, we make the following contributions:
(1) we propose a visual frontend pre-training method to correlate lip shapes with the syllabic HMM states. It does not require extra labeled/unlabeled data sets or manually-labeled word boundaries but is able to effectively utilize the visual modality; (2) we propose an audio-dominated cross-modal fusion Encoder (CMFE), in which multiple cross-modal fusions occur at different layers; (3) as a result, our AVSR system achieves a new state-of-the-art performance on the MISP2021-AVSR corpus without using extra training data and complex front-ends and back-ends.

\section{Proposed Techniques}
\subsection{Overall Training Framework}
As illustrated in Fig.~\ref{fig:framework}, our AVSR system is trained in two stages: uni-modal pre-training and multi-modal fine-tuning. In the first stage, we pre-train a hybrid audio-only ASR CTC/Attention model~\cite{watanabe2017hybrid} based on standard end-to-end ASR training as shown in the rightmost branch of Fig.~\ref{fig:framework}. For the video modality as shown in the leftmost branch of Fig.~\ref{fig:framework}, we explore a correlation between lip shapes and subword units as described in Section \ref{sec:visual frontend} to pre-train the video-only model. Then we initialize and fine-tune the audio-visual fusion model, as shown in the middle branch of Fig.~\ref{fig:framework}, after the two unit-modal networks have converged. In Fig.~\ref{fig:framework} the audio-visual fusion model integrates the audio branch (blue blocks) and visual branch (green blocks) with cross-attention blocks (red blocks). The four dashed-border blocks in the middle fusion block are initialized by the pre-trained models while the solid-border blocks are initialized randomly. Both audio-only and fusion models integrate the CTC decoder with a transformer-based decoder for joint training and decoding. The loss function can be formulated as a linear combination of the logarithm of the CTC and attention posterior probabilities as shown below:
\begin{equation}
\label{eq: loss}
\mathcal{L}_{\mathrm{MTL}}=\lambda \log P_{\text {ctc }}(Y \mid X)+(1-\lambda) \log P_{\text {att }}(Y \mid X)
\end{equation}
where $X=\left[x_1, \cdots, x_T\right]$ and $Y=\left[y_1, \cdots, y_L\right] $  denote the encoder output and the target sequences, respectively. $T$ and $L$ denote their lengths and $\lambda$ is the weight factor between the CTC loss and the attention cross entropy (CE) loss.

\subsection{Visual Pre-training by Correlating Lip Shapes with Syllable Units in Mandarin}
\label{sec:visual frontend}
In previous studies, some researchers applied the cold fusion method that freezes the pre-trained visual-only model and directly combines visual embedding with audio embedding. Others commonly pre-trained the visual frontend on the isolated word classification task and fine-tuned it with the fusion model. Compared to these techniques, our visual pre-training method correlates lip shapes with frame-level syllabic sequences generated by a GMM-HMM. It offers explicit boundaries to establish a direct frame-level mapping from lip shapes to acoustic subwords and does not need extra data sets or manually-labeled word boundaries.

\begin{figure}[htbp]
    \centering
    \hspace{-2mm}
    \includegraphics[width=1.0\columnwidth]{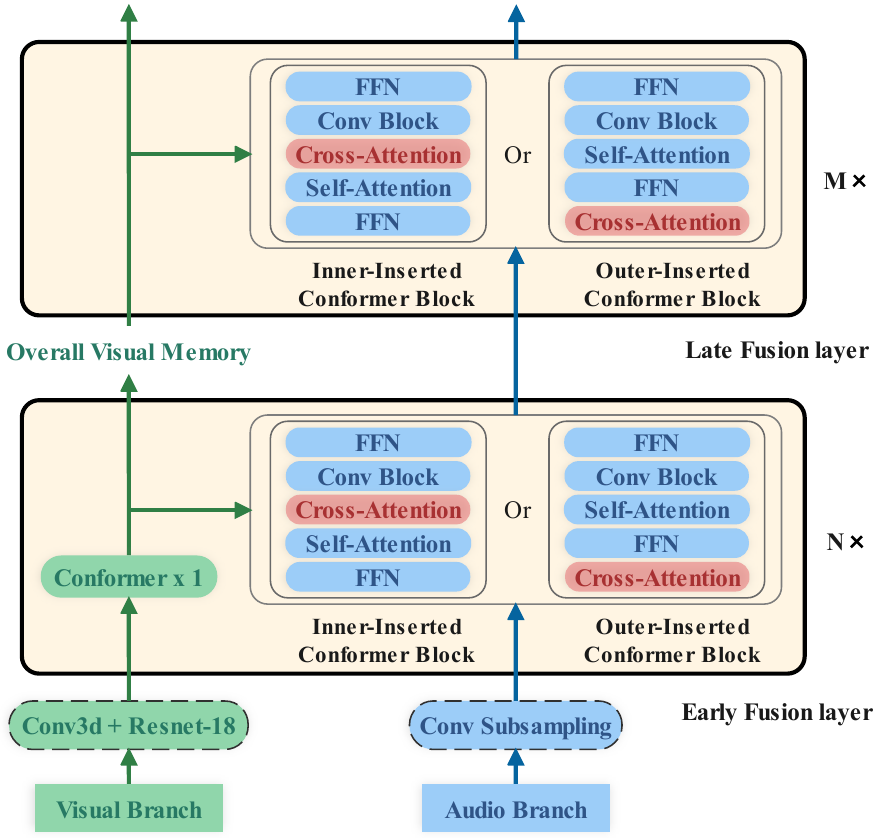}
    \caption{Cross-modal fusion encoder.}
    \label{fig:encoder}
\end{figure}
Specifically, we utilize Dacidian Dictionary~\cite{du2018aishell} as basic pronunciations to map Chinese characters to Pinyin-based syllables with tones. Then we follow the Kaldi AIShell recipe to train a triphone GMM-HMM model on far-field audio. Next, an HMM-based Viterbi forced alignment is applied to obtain the frame-level state boundaries of the clustered HMM states.
As shown on the leftmost branch of Fig.~\ref{fig:framework}, the pre-trained video-only model consists of a conv3d+resnet18 block, an up-sampling block and a 3-layer conformer block. The conv3d+resnet18 block has the same architecture as~\cite{chen2021correlating}. Due to the mismatch in sampling rate between the video frames (25fps) and alignments (100fps), two deconvolution layers are adopted to up-sample video embedding by four times. We avoid sub-sampling the tied-triphone state alignments because it could destroy a complete HMM transition. Moreover, in video recordings, a sampling process for continuous lip movement could be considered as a naturally masked operation that drops extra visual information between two frames. The up-sampling operation is intended to reconstruct dropped video frames, which helps explore potential temporal associations between two consecutive video frames.

We pre-train the video-only model using the frame-level clustered HMM state boundaries obtained in the forced alignment process described earlier. A CE criterion $\mathcal{L}_{\mathrm{CE}}$  between the output prediction posterior $P\left(Y \mid {X}\right)$ and the ground truth posterior of state $P^{\mathrm{GT}}$ is computed as follows:
\begin{equation}
\mathcal{L}_{\mathrm{CE}}=-\sum_{t=0}^{T-1} P_t^{\text{GT}} \log P\left(y_t \mid {X}\right)
\end{equation}
where $T$ is the length of the ground truth and $y_t$ is the posterior probability of corresponding HMM state classification on the $t$th frame.

\subsection{Cross-Modal Fusion Encoder}
\label{sec:encoder}
 Attention-based fusion has shown its advantages in recent studies~\cite{zhou2019modality,lee2020audio}. Unlike direct concatenation, attention-based fusion is not constrained to frame rate discrepancies and audio-visual asynchrony. Most attention-based fusion network follows a symmetric dual-branch structure without considering a modality priority. Within the two-stream framework as shown in Fig.~\ref{fig:framework}, the pre-trained audio/visual branches are already able to extract uni-modal representations at the beginning of the fusion stage, so more learning parameters are considered for modal fusion rather than uni-modal representation learning. Since speech contains more linguistic and semantic information, we reduce the depth of the visual branch, and more parameters are used for multiple modal cross-attention in different layers to make full use of modality complementarity. As a result, we degenerate the classical dual-branch structure into the audio-dominate cross-modal fusion encoder (CMFE).

As shown in Fig.~\ref{fig:encoder}, the backbone of the CMFE is composed of $N$ early fusion layers and $M$ late fusion layers with $N+M=12$ and $N \in [1,2,3] $. Compared to conventional symmetric dual-branch structures, only the early fusion layer includes a conformer block in the visual branch in our fusion model. In each fusion layer, one cross-attention block is inner/outer inserted into the conformer of the audio branch. Following the design of the decoder block of vanilla transformer, inner insertion means inserting the cross-attention between the self-attention block and the convolution block of the conformer block. Outer insertion means inserting a cross-attention layer in the front of the conformer block, which does not break the structure of a complete conformer block. For the $n^{\text{th}}$ early fusion layer, video embeddings produced by the conformer block $X_{V}^{n}$ are considered as a query (Q) to conduct cross-attention operation with audio embedding $X_{A}^{n}$ as a key (K) and a value (V) in the same layer. We consider the visual modality as the query, intended to use its robustness against acoustic signal corruption, to match target the audio components in noises. After the $n^{\text{th}}$ fusion layer, all video embedding elements from each early fusion layer are concatenated over the channel dimension and projected into a 512-dimension overall visual memory $X_{V}^{O}$ for late fusion, as formulated in the following:
\begin{equation}
X_{V}^{O} = \text{FC}(\text{Concat}(X_{V}^{1},\cdots,X_{V}^{N}))
\end{equation}
Motivated by the decoder architecture of the vanilla transformer, the same overall visual memory is directly integrated into the audio in the late fusion layers. This multiple-fusion design is aimed at reducing forgetfulness and making full use of modality complementarity in different layers.

\section{Experiments and Result Analysis}
\subsection{Experimental Setup}
\subsubsection{Data Sets and Preprocessing}
\label{subsec:preprocessing}
Most experiments are evaluated on the updated version of the MISP2021-AVSR corpus~\cite{chen2022audio}, denoted as $\mathit{MISP_{update}}$. For fairness, we experiment with the original version of the MISP2021-AVSR corpus~\cite{chen2022first} released in the MISP2021 Challenge, denoted as $\mathit{MISP_{original}}$, to compare our proposed system with the state-of-the-art systems. These two data sets share the same train set, while $\mathit{MISP_{update}}$ adds 10 hours of new data to the evaluation set to increase the data diversity. We use far-field audio and far/middle-field videos in the training stage and evaluate systems' performance on far-field audio and videos. Conventional signal processing algorithms, including weighted prediction error (WPE)~\cite{drude2018nara} and guided source separation (GSS)~\cite{boeddecker18_chime}, are applied on far-field audio for dereverberation and source separation. Then 80-dimensional filterbank feature vectors are extracted and utterance normalization is applied. We adopt speed perturbation, SpecAug~\cite{Park2019} and continuous segments splicing for data augmentation when training ASR model and only use SpecAug for AVSR. For video, we follow~\cite{chen2022audio} to obtain gray-scale lip ROI with 88$\times$88 pixels.
\renewcommand{\arraystretch}{1.2}
\begin{table}[]
\caption{Comparison of different pre-training methods using two streams. Isolated denotes Isolated word recognition and Continuous means continuous lipreading recognition.}
\label{table:framework}
\setlength{\tabcolsep}{10.15pt}{
\begin{tabular}{c|c|c|c}
\hline
Model & Pre-training Method             & Unit(number)     & CER(in \%)     \\ \hline
A0    & No Pre-training                   & \textbackslash{} & 35.53          \\
AV1   & No Pre-training                  & \textbackslash{} & 37.81          \\
AV2   & Only Pre-train A                    & \textbackslash{} & 34.97          \\
\hline
AV3   & A\&V(Isolated)& Word(500)        & 34.49          \\
AV4   & A\&V(Isolated)& Word(1000)       & 29.84          \\
AV5   & A\&V(Continuous) & Char(3385)       & 29.22          \\
AV6   & A\&V(Continuous) & Char(3385)       & 30.13          \\
AV7   & A\&V(Proposed)                  & Senone(3168)     & \textbf{28.66} \\
AV8   & A\&V(Proposed)                  & Senone(6272)     & 28.90          \\ \hline
\end{tabular}}
\vspace{-1.0em}
\end{table}

\subsubsection{Implementation Detail}
All conformers in Fig.~\ref{fig:framework} use the same set of hyper-parameters ($n_{head}$ = 8, $d_{model}$ = 512, $d_\mathit{ffn}$ = 2048, $\mathit{CNN}_{kernel}$ = 5). The attention decoder branch of the audio-only and audio-visual fusion models in Fig.~\ref{fig:encoder} consists of the 6-layer transformer ($n_{head}$ = 8, $d_{model}$ = 512, $d_\mathit{ffn}$ = 2048). We train audio-only and audio-visual fusion models with a joint CTC loss weight of $\lambda$ = 0.3. All models are optimized using Adam with $\beta_1 = 0.9$, $\beta_2=0.999$ and the learning rate of $6.0\times10^{\--4}$. The learning rate is warmed up linearly in the first 6000 steps and decreases proportionally to the inverse square root of the step number. A 6-layer transformer-based language model trained on the transcription of the training set is applied when decoding with a weight of $0.2$.

\subsection{Experiment Results}

\subsubsection{Comparison of Visual Frontend Pre-training}
In Table \ref{table:framework}, we investigate the performance of different pre-training methods following the decoupled training framework with far-field audio and far+middle field videos. Character error rate (CER) is used for all evaluations. We first train the audio-only network A0 and audio-visual fusion network AV1 from scratch, resulting in CERs of 35.53\% and 37.81\% respectively, which confirms the performance degradation method in subsection \ref{sec:intro}. We initialize the audio branch of AV2-AV8 with A0 and initialize the visual frontend of AV3-AV8 with different pre-training methods. For AV1-AV8, they have the same architecture of visual frontend as shown in the middle of Fig.~\ref{fig:framework}. A tendency is observed that audio branch initialization inhibits performance degradation, and audio-visual branch initialization achieves further performance improvement, which implies that decoupled training framework effectively mitigates variations in learning dynamics between modalities.

\begin{figure}[!t]
    \centering
    \includegraphics[width=1.0\columnwidth]{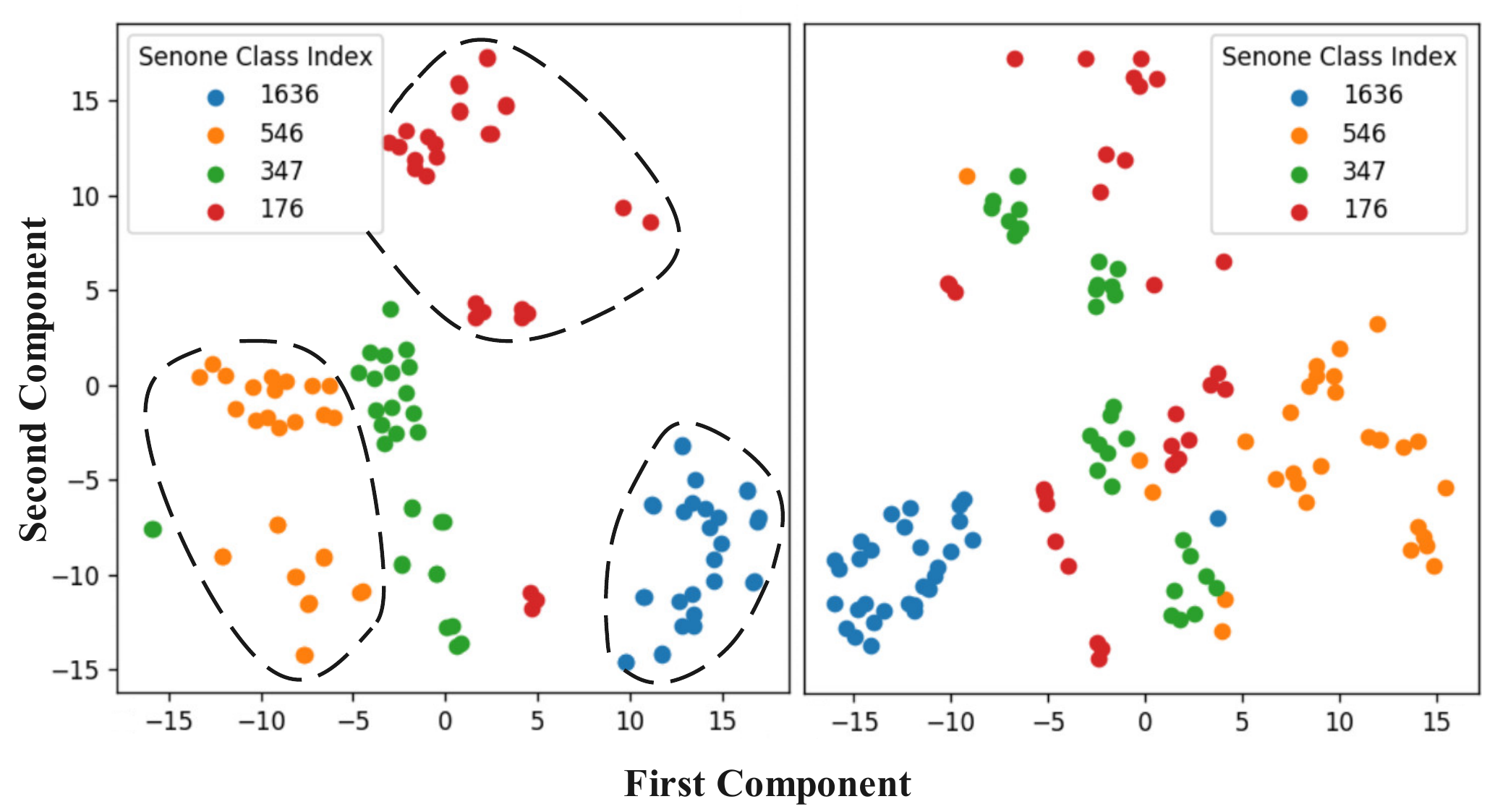}
    \caption{Resnet output embedding projection through t-SNE from AV7 (left) and AV5 (right) }
    \label{fig:scatter}
    \vspace{-1.0em}
\end{figure}

Next, we compare different visual branch pre-training methods with our proposed method. We utilize the conv3d+resnet18 modules pre-trained on isolated word recognition tasks with LRW~\cite{chung2016lip} and LRW-1000~\cite{yang2019lrw} offered by~\cite{feng2021efficient} as the visual frontend of AV3 and AV4. AV4 performs better than AV3 because LRW-1000 is a Mandarin corpus and the visual frontend of AV4 can be easily adapted to our evaluation task in MISP2021 challenge. For AV5 and AV6, their visual frontends are pre-trained on an end-to-end continuous lipreading recognition task with a CTC loss and an attention loss, respectively. Specifically, their video-only models have a similar encoder architecture to the one shown in Fig.~\ref{fig:framework} on the leftmost branch but without the up-sampling block. The video-only decoder of AV6 consists of a 6-layer transformer.
\renewcommand{\arraystretch}{1.2}
\renewcommand{\thetable}{3} 
\begin{table*}[!t]
\centering
\caption{Comparison of training data, frontend and backend to SOTA systems on $\mathit{MISP_{original}}$.}
\label{table:sota}
\setlength{\tabcolsep}{15.15pt}{
\begin{tabular}{c|cc|c|c|c}
\hline
\multirow{2}{*}{System} & \multicolumn{2}{c|}{Training Data}               & \multirow{2}{*}{Frontend} & \multirow{2}{*}{Backend Encoder} & \multirow{2}{*}{CER(in \%)} \\ \cline{2-3}
                        & \multicolumn{1}{c|}{A}          & V              &                            &                           &                             \\ \hline
NIO                     & \multicolumn{1}{c|}{3300 hours} & LRW-1000       & WPE+GSS                    & CAE (Seven Stream)         & 25.07                       \\
XIAOMI                 & \multicolumn{1}{c|}{3000 hours} & LRW-1000       & WPE+GSS+SPEx+              & AV-Encoder (Dual Stream)   & 27.17                       \\
Proposed                    & \multicolumn{1}{c|}{500 hours}  & w/o extra data & WPE+GSS                    & CMFE (Dual Stream)          & \textbf{24.58}                 \\ \hline
\end{tabular}}
\vspace{-1.0em}
\end{table*}
\smallskip
\renewcommand{\arraystretch}{1.2}
\renewcommand{\thetable}{2} 
\begin{table}[!t]
\caption{Comparison of different audio-visual modal fusion strategies.}
\label{table:encoder}
\centering
\setlength{\tabcolsep}{10.15pt}{
\begin{tabular}{c|c|c|c|c}
\hline
Model & Fusion  & $P_{insert}$ & $N_{vblock}$ & CER(in \%) \\ \hline
AV9    & TM-CTC & Outer   & 3        & 28.98      \\
AV10    & TM-Seq & Outer   & 3        & 34.70      \\
AV7    & Baseline & Outer   & 3        & 28.66 \\
\hline

AV11    & CMFE      & Outer   & 3        & 28.00      \\
AV12    & CMFE      & Outer   & 2        & \textbf{27.90}       \\
AV13   & CMFE      & Outer   & 1        & 28.13      \\ \hline
AV14    & CMFE      & Inner   & 3        & 28.32      \\
AV15    & CMFE      & Inner   & 2        & 28.09      \\
AV16    & CMFE      & Inner   & 1        & 28.15      \\ \hline
\end{tabular}}
\vspace{-2em}
\end{table}
As shown in Table~\ref{table:framework}, the visual front-ends of AV7 and AV8 are pre-trained on our proposed method. The only difference is the number of clustered HMM states. AV7 and AV8 perform better than other pre-training methods in AV3-AV6, and AV7 model trained with fewer senone units achieves a slightly better CER than AV8 (28.66\% vs. 28.90\%). The results show an advantage of the fine-grained alignment labels that offer frame-level syllable boundaries to guide visual feature extraction. Moreover, we use t-distributed Stochastic Neighbor Embedding to visualize the output embedding from the pre-trained visual frontend of AV5 and AV7 in Fig~\ref{fig:scatter} and observe that the embedding projection of AV7 shows a better clustering on syllable units than AV5. It indicates that using the fine-grained frame-level syllable labels enables the visual frontend to explore potential acoustic information from lip movements and contributes to a better adaptation with the audio stream in the fusion stage.

\subsubsection{Comparison of Audio-visual Fusion Strategies}
The results of different fusion strategies on $\mathit{MISP_{update}}$ are shown in Table~\ref{table:encoder}. All models are trained following the decoupled training framework. $P_{insert}$ denotes that the cross-attention block is inner/outer inserted in the conformer block as shown in Fig.~\ref{fig:encoder}. $N_{vblock}$ is the number of conformer blocks in the visual branch of the fusion model. TM-CTC~\cite{afouras2018deep} and TM-Seq~\cite{afouras2018deep} are two classic attention-based fusion structures in which audio and visual streams are integrated in the encoder/decoder respectively. The architecture of the Baseline is shown in the middle branch of Fig.~\ref{fig:framework}. Compared with TM-CTC, TM-Seq and Baseline, the proposed CMFE (AV11-AV16) performs better with multiple fusions in different layers. And the best model (AV12) achieves an CER of 27.90\% on $\mathit{MISP_{update}}$ (more difficult than $\mathit{MISP_{original}}$ with an CER of 26.21\%). We gradually decrease the number of conformer blocks in the visual branch, and no obvious performance drop is observed. It indicates that more training parameters can be used for modality fusion within the decoupled training framework attributed to visual frontend initialization. Finally, we compare two methods of inserting cross-attention blocks into the original conformer layer. Outer insertion (AV11-AV13) slightly outperforms inner insertion (AV14-AV16) as it does not break the complete conformer block structure.

\subsubsection{Overall Comparison with State-of-the-art Systems}
In Table~\ref{table:sota} we present an overall comparison of our proposed system, NIO system~\cite{xu2022channel} and XIAOMI system~\cite{wangxiaomi} (the 1st and 2nd place in MISP2021 Challenge). We compare these systems in terms of audio-visual training data, frontend, and backend encoders. For training, NIO and XIAOMI adopted all far/middle/near-field audio and applied a series of simulation and augmentation methods to extend the training set to 3300 and 3000 hours, respectively. Both of them initialized their visual branch on isolated word tasks with extra word-level data sets, LRW-1000. In comparison, our audio-only model is trained on a 500-hour data set and we do not use any extra data to pre-train the visual branch. In terms of frontend and backend, XIAOMI trained an extra neural signal separator (SPEx+~\cite{ge2020spex+}) for source separation. NIO had a large-size backend encoder denoted as CAE to process all original 6-channel signals, the enhanced channel, and the visual features. In contrast, our system is simple yet effective with the frontend system consisting of GSS and WPE and a dual-stream backend. We then use the recognizer output voting error reduction (ROVER)~\cite{fiscus1997post} procedure to rescore the output transcripts of A0, AV7, AV12, AV15 models in Tables~\ref{table:framework}-\ref{table:encoder}. As a result, our system attains a state-of-the-art CER of 24.58\% and outperforms the NIO system ~\cite{xu2022channel} by an absolute CER reduction of 0.5\%. On the more difficult $\mathit{MISP_{update}}$ test, our proposed ROVER system also gives a good CER of 25.96\%.

\section{Conclusion}
In this paper, we decouple one-pass end-to-end AVSR training into two stages to mitigate modality variations. Furthermore, we propose a visual pre-training framework by correlating lip shapes with syllables to establish good frame-level syllable boundaries
from lip shapes. Moreover, a novel CMFE block is introduced to model multiple cross-modal attentions in the fusion stage and
 make full use of multi-modal complementarities. Compared to the currently top-performance systems in MISP2021-AVSP Challenge, our proposed system is simple yet effective and achieves a new state-of-the-art performance without using extra training data
and complex front-ends and back-ends. In the future, more types of subword units, such as visemes and phonemes, will be explored to improve correlation-based visual pre-training and cross-modal fusion encoder.

\section*{Acknowledgment}
This work was supported by the National Natural Science Foundation of China under Grant No. 62171427 and the Strategic Priority Research Program of Chinese Academy of Sciences under Grant No.XDC08050200.
\balance
\vfill\pagebreak
\bibliography{main}
\bibliographystyle{IEEEbib}

\end{document}